\documentclass[a4paper, 10pt, conference]{ieeeconf} 

\IEEEoverridecommandlockouts                             
\usepackage{graphicx}
\usepackage{amsmath}
\usepackage{amssymb}
\usepackage{booktabs}
\usepackage{subcaption}

\usepackage[binary-units=true]{siunitx}
\DeclareSIUnit\px{px}
\DeclareSIUnit{\nothing}{\relax}

\usepackage[mode=buildnew]{standalone}
\usepackage{scalerel}
\usepackage{rotating}

\usepackage{xcolor}
\usepackage{tikz}

\usepackage[pagebackref,breaklinks,colorlinks]{hyperref}

\usepackage[capitalize]{cleveref}
\crefname{section}{Sec.}{Secs.}
\Crefname{section}{Section}{Sections}
\Crefname{table}{Table}{Tables}
\crefname{table}{Tab.}{Tabs.}

\usepackage{caption}
\usepackage{mathrsfs} 
\usepackage{xspace}
\usepackage{multirow}
\usepackage{bbding}
\usepackage{pifont}
\newcommand{\dataset}{TAS-NIR dataset\xspace}
\newcommand{\numtasnir}{209\xspace}
\newcommand{\numtrain}{439\xspace}

\DeclareMathOperator*{\argmin}{arg\,min}
\DeclareMathOperator*{\ndvi}{\text{NDVI}}
\DeclareMathOperator*{\nir}{\text{NIR}}
\DeclareMathOperator*{\red}{\text{R}_{\mathrm{VIS}}}
\DeclareMathOperator*{\evi}{\text{EVI}}
\DeclareMathOperator*{\blue}{\text{B}_{\mathrm{VIS}}}

\newcommand{\xmark}{\ding{55}}%

\definecolor{viscam}{RGB}{69, 161, 217}
\definecolor{nircam}{RGB}{237, 110, 0}
\definecolor{linegray}{RGB}{180,180,180}
\definecolor{asphalt}{RGB}{ 192,192,192}
\definecolor{gravel}{RGB}{ 105,105,105}
\definecolor{soil}{RGB}{ 160, 82, 45}
\definecolor{sand}{RGB}{ 244,164, 96}
\definecolor{bush}{RGB}{  60,179,113}
\definecolor{forest}{RGB}{  34,139, 34}
\definecolor{low_grass}{RGB}{ 154,205, 50}
\definecolor{high_grass}{RGB}{   0,128,  0}
\definecolor{scenery_vegetation}{RGB}{   0,100,  0}
\definecolor{tree_crown}{RGB}{   0,250,154}
\definecolor{tree_trunk}{RGB}{ 139, 69, 19}
\definecolor{building}{RGB}{   1, 51, 73}
\definecolor{fence}{RGB}{ 190,153,153}
\definecolor{wall}{RGB}{   0,132,111}
\definecolor{car}{RGB}{   0,  0,142}
\definecolor{bus}{RGB}{   0, 60,100}
\definecolor{sky}{RGB}{ 135,206,250}
\definecolor{obstacle}{RGB}{ 128,  0,128}
\definecolor{pole}{RGB}{ 153,153,153}
\definecolor{traffic_sign}{RGB}{ 255,255,  0}
\definecolor{person}{RGB}{ 220, 20, 60}
\definecolor{animal}{RGB}{ 255,182,193}
\definecolor{ego_vehicle}{RGB}{ 220,220,220}
\definecolor{undefined}{RGB}{   0,  0,  0}

\begin{document}

\title{\LARGE \bf TAS-NIR: A VIS+NIR Dataset for Fine-grained Semantic Segmentation in Unstructured Outdoor Environments \\
	\thanks{The authors gratefully acknowledge funding by the Federal Office of Bundeswehr Equipment, Information Technology and In-Service Support (BAAINBw).}}

\author{Peter Mortimer\\
Universität der Bundeswehr München\\
Institute for Autonomous Systems Technology\\
{\tt\small peter.mortimer@unibw.de}
\and
Hans-Joachim Wünsche\\
Universität der Bundeswehr München\\
Institute for Autonomous Systems Technology\\
{\tt\small joe.wuensche@unibw.de}
}
\maketitle

\begin{abstract}
   Vegetation Indices based on paired images of the visible color spectrum (VIS) and near infrared spectrum (NIR) have been widely used in remote sensing applications. 
   These vegetation indices are extended for their application in autonomous driving in unstructured outdoor environments. 
   In this domain we can combine traditional vegetation indices like the Normalized Difference Vegetation Index (NDVI) and Enhanced Vegetation Index (EVI) with Convolutional Neural Networks (CNNs) pre-trained on available VIS datasets. 
   By laying a focus on learning calibrated CNN outputs, we can provide an approach to fuse known hand-crafted image features with CNN predictions for different domains as well. 
   The method is evaluated on a VIS+NIR dataset of semantically annotated images in unstructured outdoor environments. 
   The dataset is available at \href{https://mucar3.de/iros2022-ppniv-tas-nir/}{\textcolor{black}{mucar3.de/iros2022-ppniv-tas-nir}}.
\end{abstract}

\section{Introduction}
\label{sec:intro}

Most computer vision approaches to autonomous driving consider the visible spectrum (VIS) during image acquisition. 
This led to many large-scale semantic segmentation datasets for urban driving scenarios.
Only few annotated datasets have been released for unstructured outdoor driving scenarios \cite{pemo:rugd_wigness2019iros,pemo:rellis3d_jiang2020arxiv,pemo:tas500_metzger2020icpr,pemo:ycr_maturana2018fsr} and even fewer consider image data beyond the visible spectrum \cite{pemo:freiburgforest_valada2017adapnet}. \\
Foliage has a high reflectivity in the near infrared (NIR) spectrum, which is also known as the Wood effect \cite{pemo:woodeffect_wood1910phj}. Partly based on this observation, a few vegetation indices using visible light and near infrared light (VIS+NIR) were developed for Remote Sensing. This motivates the use of the VIS+NIR spectrum for vegetation and ground surface segmentation in autonomous driving in unstructured outdoor environments. 
The lack of large training datasets of VIS+NIR image pairs prevent end-to-end deep learning approaches, especially for a fine-grained semantic segmentation. \\
To alleviate the imbalance of semantically segmented VIS images and NIR images, we suggest a late fusion of predictions made by neural networks pre-trained on larger VIS datasets with predictions coming from hand-crafted vegetation indices based on smaller VIS+NIR datasets.
Additionally, the VIS image prediction should produce calibrated outputs to more accurately resemble the confidence of a given class prediction. 
\begin{figure}[t]
	\centering
	\includegraphics[width=.93\linewidth]{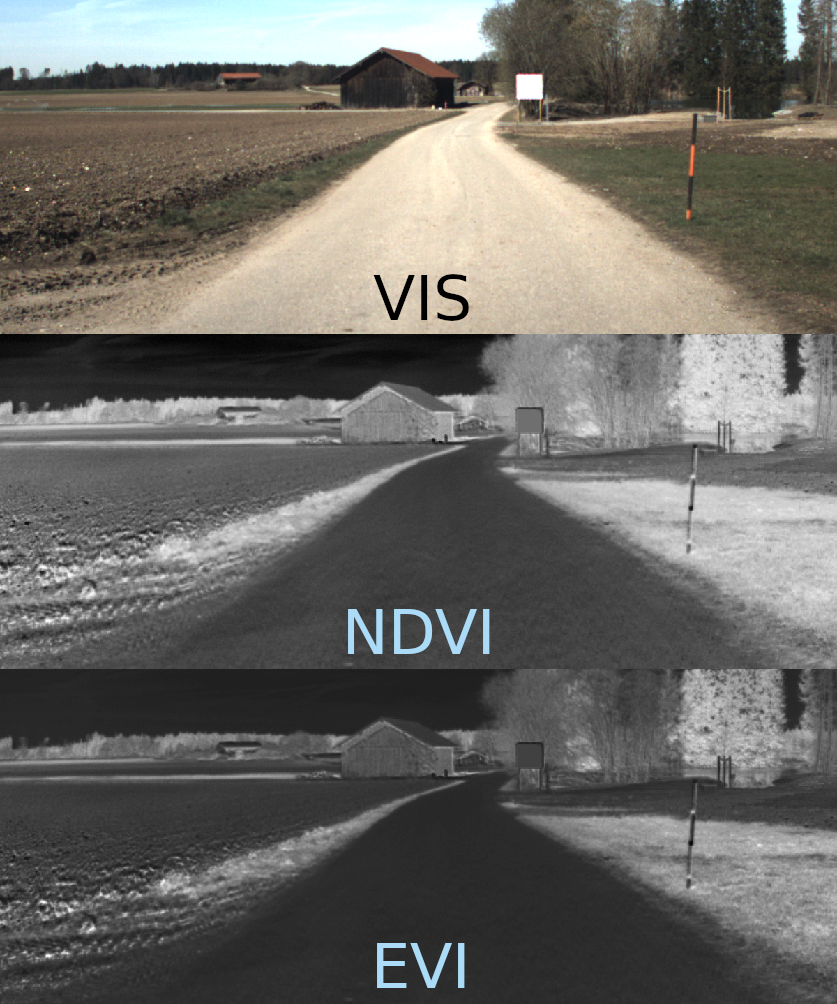}
	\caption{For the fine-grained semantic segmentation of ground surfaces and vegetation we leverage both calibrated neural networks using images from the visible spectrum (top) and hand-crafted vegetation indices like the Normalized Difference Vegetation Index (middle) and the Enhanced Vegetation Index (bottom).}
	\label{fig:title-card}
\end{figure}
\begin{figure*}[htpb!]
	\centering
	\begin{subfigure}{0.37\linewidth}
		\includegraphics[width=\linewidth]{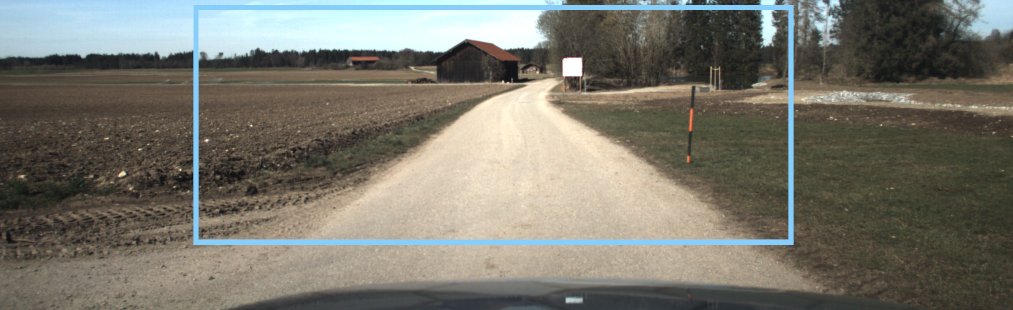}
		\caption{The \textcolor[RGB]{135, 206, 250}{ROI} from the VIS image.}
		\label{fig:projection-vis}
	\end{subfigure}
	\hfill
	\begin{subfigure}{0.24\linewidth}
		\includegraphics[width=\linewidth]{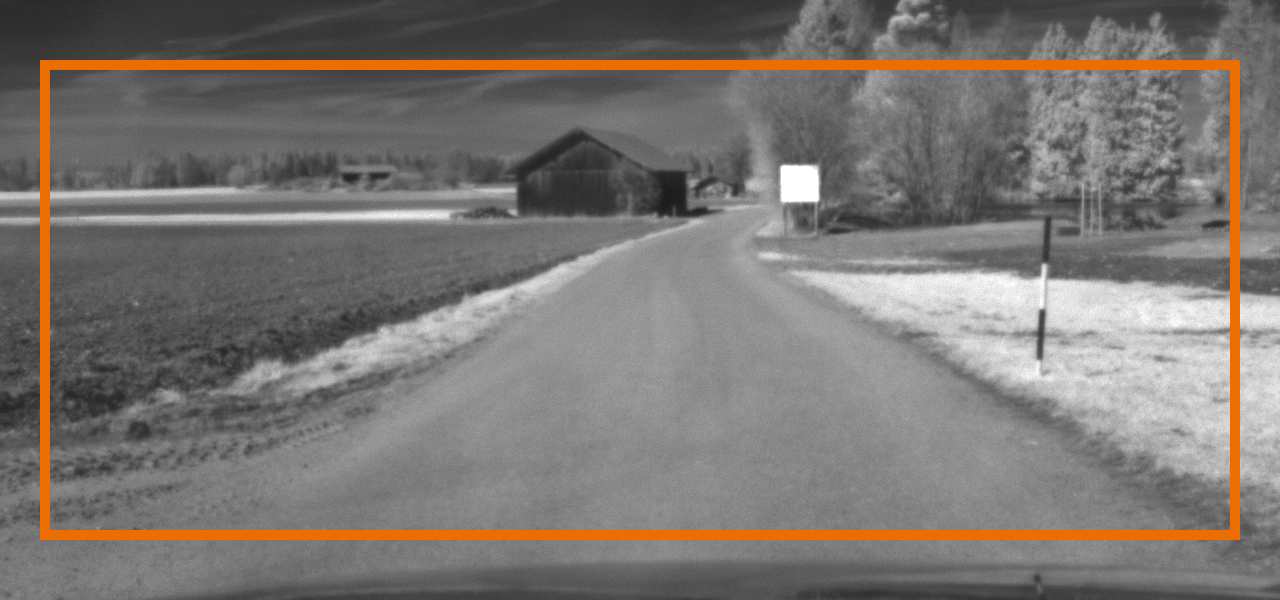}
		\caption{The \textcolor[RGB]{237, 110, 0}{ROI} from the NIR image.}
		\label{fig:projection-nir}
	\end{subfigure}
	\hfill
	\begin{subfigure}{0.37\linewidth}
		\includegraphics[width=\linewidth]{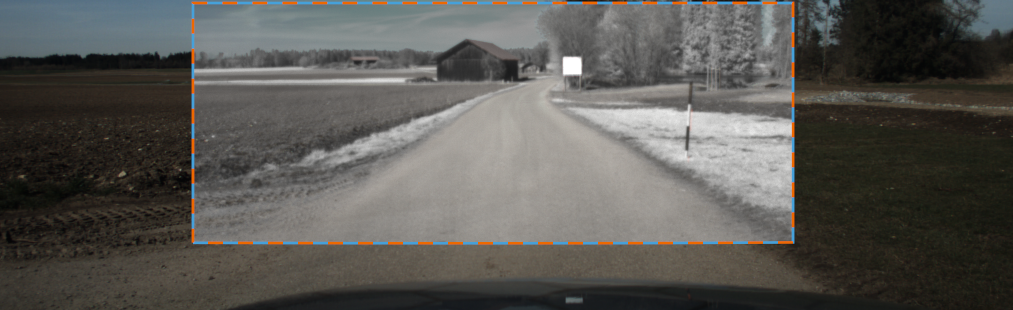}
		\caption{\textcolor[RGB]{135, 206, 250}{VIS}+\textcolor[RGB]{237, 110, 0}{NIR} after homography transform $H_{\scaleto{\mathrm{NIR} \to \mathrm{VIS}}{3pt}}$.}
		\label{fig:projection-visnir}
	\end{subfigure}
	\caption{The overlapping regions of interest (ROI) are matched, and a homography matrix $H_{\scaleto{\mathrm{NIR} \to \mathrm{VIS}}{3pt}}$ is applied onto the NIR image to also match the perspective of the VIS image (see \cref{eq:homography}). The perspective transform based on a ground surface homography can lead to pixel mismatches especially along the border of close obstacles like the pole in \cref{fig:projection-visnir}.}
	\label{fig:projection-images}
\end{figure*}
The performance improvement of our approach will be quantitatively compared with previous approaches on a fine-grained semantic segmentation VIS+NIR dataset which we release as part of this publication. 
We make the following contributions in this paper:
\begin{itemize}
	\item[-] A method for the late fusion of calibrated neural network predictions on VIS images and hand-crafted  VIS+NIR features like NDVI and EVI. We apply this for the fine-grained semantic segmentation of ground surfaces and vegetation types in unstructured outdoor environments. 
	\item[-] A novel dataset consisting of \numtasnir semantically segmented and aligned VIS+NIR images in different driving scenarios in unstructured outdoor environments. 
	The fine-grained semantic segmentation of the different vegetation and ground surface types allows closer analysis of VIS+NIR based features.
\end{itemize}


\section{Related Work}

\subsection{NIR in Computer Vision}
\noindent NIR images are used in security applications, because NIR images can reveal content similar to VIS images while not requiring a light source visible to humans \cite{pemo:nirface_li2007pami}. 
The difference of radiance between the NIR spectrum and the visible red color spectrum has been used to detect vegetation in remote sensing applications \cite{pemo:ndvi_rouse1974nasa}. 
This observation formed the basis for vegetation indices such as NDVI \cite{pemo:ndvi_kriegler1969nasa} and EVI \cite{pemo:evi_huete1999modis}. 
The common use of the VIS+NIR spectrum for vegetation detection in remote sensing motivated the use of VIS+NIR imagery for ground-level robotics in forested environments \cite{pemo:forested_valada2016iser}. 
The NIR spectrum has also been shown to enhance VIS imagery by dehazing and providing higher contrast along object boundaries \cite{pemo:dehazing_schaul2009icip,pemo:colornir_fredembach2008istsid}. 
In our work, we present the application of NIR imagery for autonotmous driving in unstructured outdoor environments.

\subsection{VIS+NIR Semantic Segmentation}
\noindent A semantic segmentation using a conditional random field (CRF) of VIS+NIR scenery images has shown an improvement in prediction accuracy over VIS images for semantic classes like \textit{sky}, \textit{vegetation} and \textit{water}, whose response in the NIR domain is discriminant \cite{pemo:scene_salamati2012eccv}. 
For unmanned ground vehicles (UGVs), neural network architectures training on multimodal images of forested environments have been developed. 
These relied on multiple independent expert networks trained on the different image modalities and included depth images as well \cite{pemo:freiburgforest_valada2017adapnet}. This is extended by self-supervised fusion mechanisms trained to adapt to the spatial location and object class in the image \cite{pemo:ssma_valada2019ijcv}. 
The more intricate fusion strategies were deemed necessary compared to a channel-wise stacking of the input modalities, where networks did not learn to leverage complementary features and cross-modal interdependencies \cite{pemo:fusenet2016accv}. 
These works segment the scene using very broad object classes like \textit{grass} and \textit{tree} for vegetation types and \textit{soil} and \textit{road} for surface types.
In our work, we make use of the NIR domain for a more fine-grained segmentation of the vegetation and surface types found in unstructured outdoor environments.

\subsection{Calibrated Neural Network Predictions}
\noindent Modern neural networks for image classification have shown to be overconfident in their prediction probability estimates, when compared to their true correctness likelihood. 
To obtain more accurate confidence scores, a so-called temperature scaling optimized over the negative log-likelihood (NLL) on the validation set can be added as a post-processing after the network's prediction \cite{pemo:tempscaling_guo2017icml}. 
Other approaches use network ensembles \cite{pemo:deepensemble_lakshminarayanan2017neurips} or Monte Carlo dropout \cite{pemo:mcdropout_gal2016icml} at test time to estimate predictive uncertainty. 
The temperature scaling, which represents a single-parameter variant of Platt Scaling \cite{pemo:plattscaling_platt1999almc}, was extended for semantic segmentation tasks by generating a temperature map in local temperature scaling (LTS) \cite{pemo:lts_ding2021iccv}.
LTS incorporates an image and location dependent temperature map to account for different miscalibration effects, such as accurate predictions for object interiors and ambiguities at near-boundary locations in semantic maps. 
Our approach relies on LTS to produce calibrated prediction outputs for our probabilistic model.

\section{Dataset}

\noindent We propose the novel \dataset to investigate the relationship of VIS+NIR images for a fine-grained vegetation and ground surface segmentation. 
The \dataset is unique due to its fine-grained semantic segmentation of the driving scenes in unstructured outdoor environments and the inclusion of NIR images in addition to the VIS images.

\begin{table*}
	\centering
	\begin{tabular}{*{5}{c|c|c|c|c}}
		Dataset & No. Scenes & Resolution & Scene Type & Annotation Type \\
		\cline{1-5}
		EPFL RGB-NIR Scene Dataset \cite{pemo:visnirscenes_brown2011cvpr} & 477 & $1024~\times$ \SI{768}{\px} & Indoor \& Outdoor Photography & Image Classification \\
		EPFL Semantic Segmentation Dataset \cite{pemo:scene_salamati2012eccv} & $370^{\dagger}$ & $1024~\times$ \SI{768}{\px} & Outdoor Photography & Semantic Segmentation \\
		HyKo2 \cite{pemo:hyko_winkens2017iccvw} & $78^{\ddagger}$ & $214~\times$ \SI{417}{\px} & Outdoor Driving & Semantic Segmentation \\
		Freiburg Forest \cite{pemo:freiburgforest_valada2017adapnet} & 366 & $1024~\times$ \SI{768}{\px} & Outdoor Driving & Semantic Segmentation \\
		\textbf{TAS-NIR (ours)} & $209$ & $1200~\times$ \SI{480}{\px} & Outdoor Driving & Fine-grained Semantic Segmentation
	\end{tabular}
	\caption{A comparison of known VIS+NIR datasets in terms of their size, scene type and annotation type. $\dagger$: The outdoor scenes from the EPFL Semantic Segmentation Dataset are only compared here. $\ddagger$: Only scenes taken with the MQ022HG NIR camera are considered in the HyKo2 dataset. 
	}
	\label{tab:vis-nir-datasets}
\end{table*}

\begin{figure}[h]
	\centering
	\includegraphics[width=.9\columnwidth, trim={10cm 0 0 0},clip]{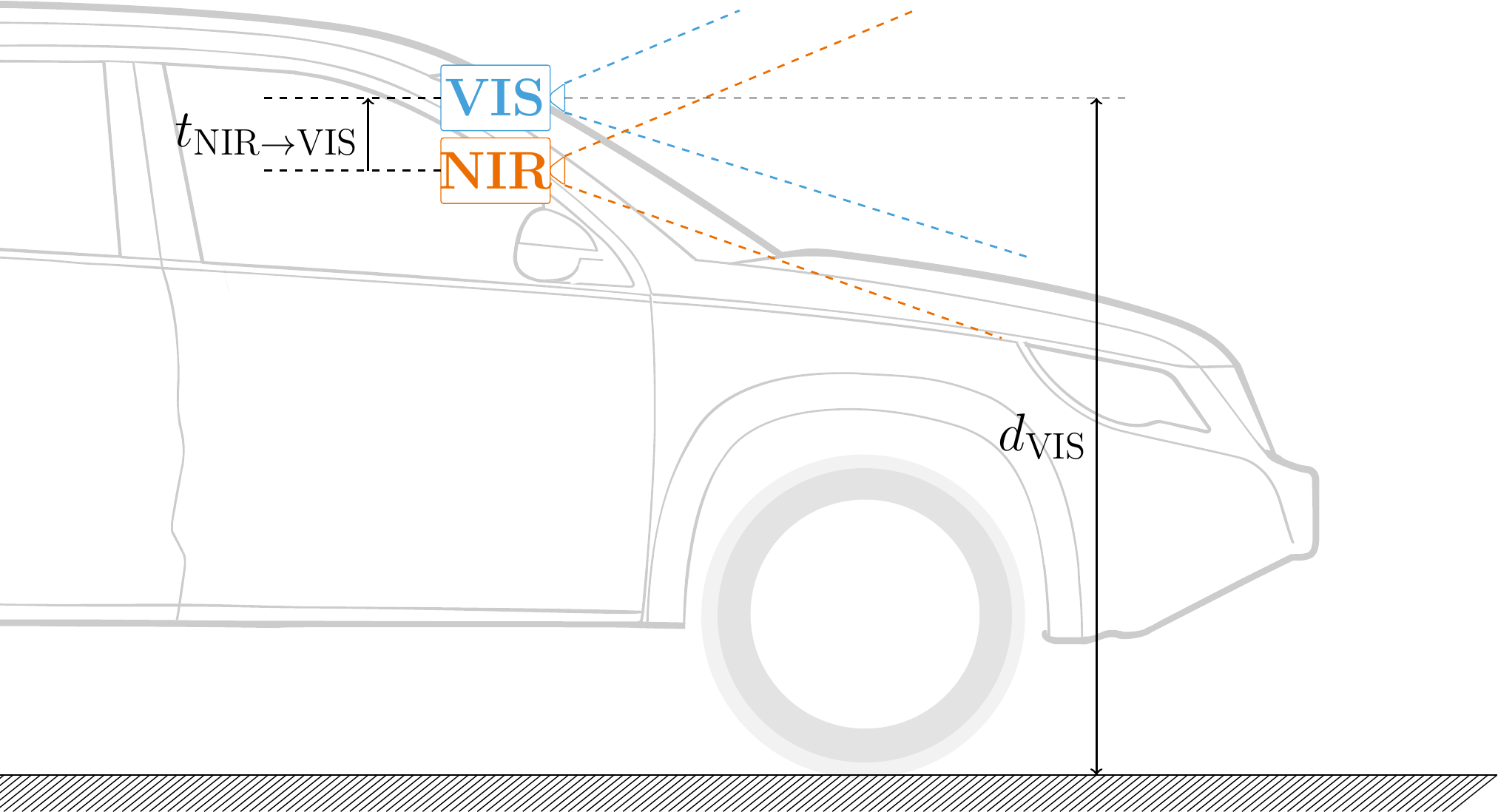}
	\caption{The NIR camera is positioned directly under the VIS camera. The shared orientation and the known height of the vehicle over a flat ground surface allows the construction of a homography matrix $H_{\scaleto{\mathrm{NIR} \to \mathrm{VIS}}{3pt}}$.}
	\label{fig:projection-graph}
\end{figure}

The dataset was recorded using our survey vehicle while driving in different unstructured outdoor environments during spring, summer and autumn.  
The VIS images and NIR images were recorded with two cameras mounted on a camera platform \cite{pemo:marveye_unterholzner2010iv}. 
The NIR camera is mounted directly under the VIS camera, and both cameras share the same orientation.
The region of interest, where both the VIS and NIR images overlap, is shown in \cref{fig:projection-images}.
To record the NIR image we use a Basler acA1300-60gmNIR with a built-in EV76C661 CMOS sensor. 
A longpass filter is attached onto the lens of the NIR camera to prevent any light under \SI{765}{\nano\meter} to pass through the lens.\\
We transform the perspective of the NIR image to match that of the VIS image by applying a homography matrix $H_{\scaleto{\mathrm{NIR} \to \mathrm{VIS}}{3pt}}$. The homography matrix $H_{\scaleto{\mathrm{NIR} \to \mathrm{VIS}}{3pt}}$ is constructed assuming a flat ground plane visible in both cameras and knowing the height of the cameras over the ground plane. 
The homography is only an approximation of the true geometry.
\begin{equation}
H_{\scaleto{\mathrm{NIR} \to \mathrm{VIS}}{3pt}} = K_{\scaleto{\mathrm{VIS}}{3pt}} \left( R_{\scaleto{\mathrm{NIR}}{3pt}} + \frac{t_{\scaleto{\mathrm{NIR} \to \mathrm{VIS}}{3pt}}~n^{T}_{\scaleto{\mathrm{NIR}}{3pt}}}{d_{\scaleto{\mathrm{NIR}}{3pt}}} \right)~K^{-1}_{\scaleto{\mathrm{NIR}}{3pt}}
\label{eq:homography}
\end{equation} 
\cref{fig:projection-graph} shows the camera setup in the vehicle. The \dataset consists of \numtasnir VIS+NIR image pairs with a fine-grained semantic segmentation. Similar to the WildDash 2 benchmark \cite{pemo:wilddash_zendel2018eccv}, the \dataset does not provide enough images to train algorithms by itself and should primarily be used for validation and testing. 
The available data from the TAS500 dataset \cite{pemo:tas500_metzger2020icpr} uses the same labeling policy, but does not provide a NIR image. The TAS500 dataset is therefore used to train a semantic segmentation model solely on VIS images. In \cref{tab:vis-nir-datasets}, we compare the \dataset with other available VIS+NIR datasets.

\section{Methodology}

\subsection{Local Temperature Scaling}
\noindent Neural networks for semantic segmentation tasks make use of the softmax function $\sigma_{\mathrm{SM}}$ on the network logits $\mathbf{z}(x)$ of each output pixel location $x$ to produce the confidence score $\hat{p}(x)$. 
The network logits $\mathbf{z}(x)$ is a vector of length $k$ with a logit for each possible semantic class.
\begin{equation}
\sigma_{\mathrm{SM}}(\mathbf{z}(x))^{(k)} = \frac{\mathrm{exp}(z(x)^{(k)})}{\sum\limits_{j=1}^{K} \mathrm{exp}(z(x)^{(j)})}
\label{eq:softmax}
\end{equation}
Traditionally, the confidence score was taken to be the largest activation from the softmax function.
\begin{equation}
\hat{p}(x) = \max_{k}\sigma_{\mathrm{SM}}(\mathbf{z}(x))^{(k)}
\label{eq:max-network-output}
\end{equation}
Neural networks, especially recent models trained with batch normalization, have shown to produce overconfident output probabilities during later stages of training \cite{pemo:tempscaling_guo2017icml,pemo:focalloss_mukhoti2020neurips}.
This effect is less of a problem for applications, where the network output is the final stage of the perception process.
For our probabilistic model, where we fuse the network output probability of a network based solely on VIS images with hand-crafted vegetation indices from VIS+NIR image pairs, we want the network output probability to resemble the empirically observable segmentation errors. 
The process of adjusting the network output probabilities is called calibration. \\
The temperature concept, which applies a Platt scaling with a single scale parameter $T > 0$ for all classes can be used to calibrate the network outputs \cite{pemo:tempscaling_guo2017icml}. 
The temperature $T$ can raise the output entropy of the network by softening the softmax function for $T > 1$. 
$T$ is optimized with respect to the NLL on the validation set to produce a calibrated confidence value $\hat{q}$.
\begin{equation}
\hat{q}(x) = \max_{k}\sigma_{\mathrm{SM}}(\mathbf{z}(x)/T)^{(k)}
\label{eq:temperature-scaling}
\end{equation}
Temperature scaling was originally intended for image classification tasks, where the prediction on the input image corresponds to a network output vector of logits. \\
For semantic segmentation tasks, where the network produces output logits $\mathbf{z(x)}$ for each pixel location $x$ in the input image, the temperature concept was extended in Local Temperature Scaling (LTS)~\cite{pemo:lts_ding2021iccv}. 
The calibration is now extended from calibrated predictions $\hat{q}(x)$ being adjusted by the same scalar temperature value $T$ to an image $n$ and location $x$ dependent probability map $\hat{\mathbb{Q}}_{n}(x, T_{n}(x))$ and temperature map $T_{n}(x)$.
\begin{equation}
\hat{\mathbb{Q}}_{n}(x, T_{n}(x)) = \max_{k \in K} \sigma_{\mathrm{SM}}\big(\mathbf{z}_{n}(x) / T_{n}(x)\big)^{(k)} 
\label{eq:temperature-map}
\end{equation} 
\begin{equation*}
\textit{where }T_{n}(x) \in \mathbb{R}^{+} \textit{is image and location dependent.}
\end{equation*}
The image and location dependence is modeled by a small neural network $\mathscr{H}$ which takes the input image $I$ and network output logits map $\mathbf{Z}$. So $\mathscr{H}$ learns the mapping $(I,\mathbf{Z}) \to \hat{\mathbb{Q}}$. $\mathscr{H}$ uses a tree-like convolutional neural network and optimizes over the NLL of the $x$ pixel locations on the $N_{val}$ images in the hold-out validation dataset \cite{pemo:autoencodertrees_irsoay2016acml, pemo:poolingtree_lee2018pami}.
\begin{figure}[h]
	\begin{subfigure}{0.495\linewidth}
		\includegraphics[width=\linewidth]{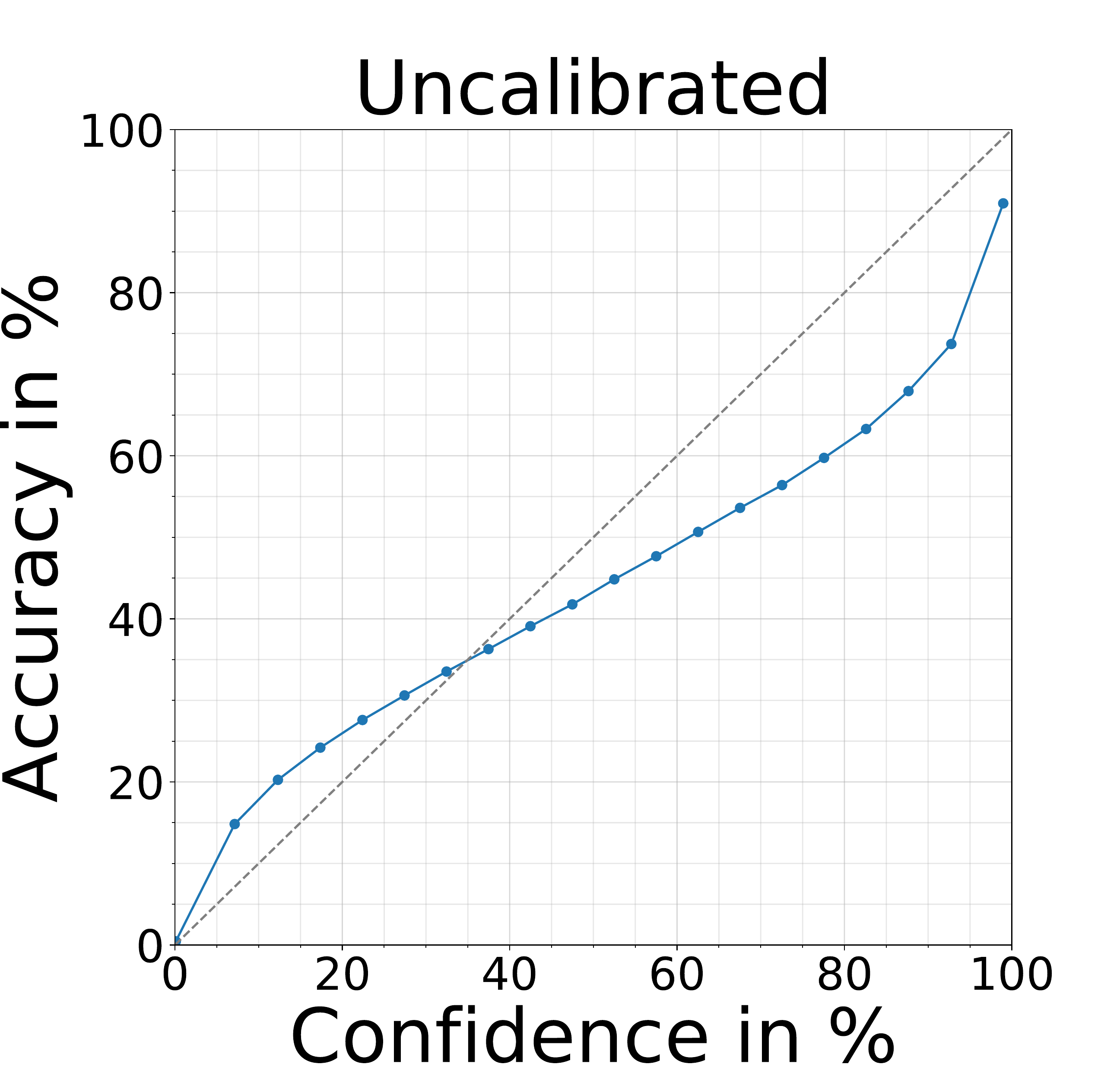}
		\label{fig:reliability-diagram-calibrated}
	\end{subfigure}
	\hfill
	\begin{subfigure}{0.495\linewidth}
		\includegraphics[width=\linewidth]{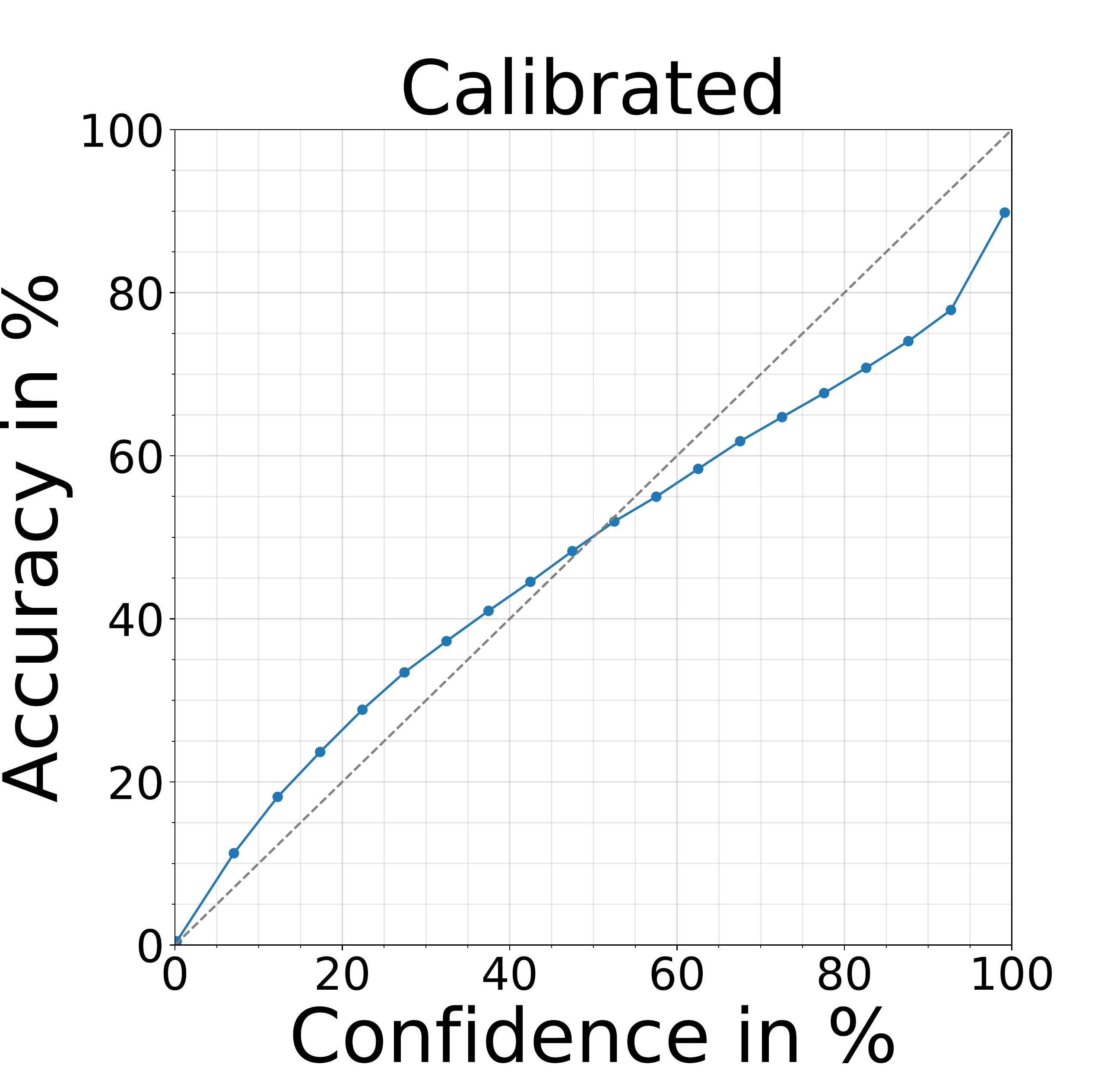}
		\label{fig:reliability-diagram-uncalibrated}
	\end{subfigure}
	\vspace{-1.5em}
	\caption{The reliability diagrams \cite{pemo:calibrationcurve_niculescu2005icml} for an uncalibrated and calibrated DeepLabV3+ network on the \dataset test split. A reliability diagram can give insight on the calibration of a predictive model. Here we compare the confidence $\hat{q}(x)$ scores for each pixel with the accuracy reported when comparing it to the ground truth of the test split. A well calibrated model would have its confidence values close to its eventual accuracy (e.g.\@ only 40\% of all pixels with a confidence score of 40\% are correctly classified). The ideal calibration is signified by the gray dashed line.}
	\label{fig:reliability-diagram}
\end{figure}
\begin{equation}
\theta^{*} = \argmin_{\theta} -\sum\limits_{n=1}^{N_{\mathrm{val}}} \sum\limits_{x \in \Omega} \log \Big( \sigma_{\mathrm{SM}} \big ( \frac{\textbf{z}_n (x)}{\mathscr{H}(\theta, \textbf{Z}_n, I_n, x)} \big ) \Big)
\label{eq:nll-optimization}
\end{equation}
\begin{equation*}
s.t. \quad \mathscr{H}(\theta, \textbf{Z}_n, I_n, x) > 0
\end{equation*}

Practically speaking, LTS adds an additional post-processing stage to the overall training process, where we optimize the weights of the temperature mapping network $\mathscr{H}$. 
This is alleviated by the few parameters in $\mathscr{H}$ and the relatively small size of the hold-out validation dataset. The effects of network output calibration can be observed in the reliability diagrams in \cref{fig:reliability-diagram}.

\subsection{Vegetation Indices}
\begin{figure}
	\begin{subfigure}{0.495\linewidth}
		\includegraphics[width=\linewidth]{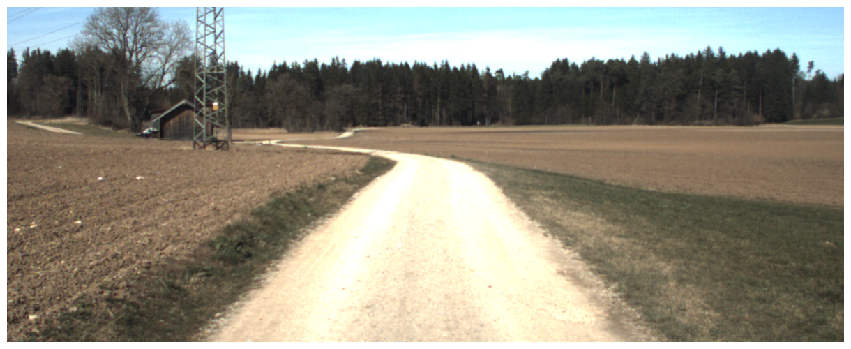}
		\caption*{VIS}
	\end{subfigure}
	\begin{subfigure}{0.495\linewidth}
		\includegraphics[width=\linewidth]{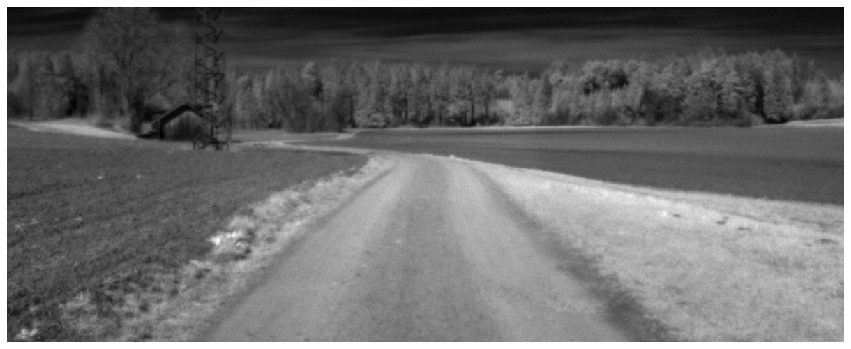}
		\caption*{NIR}
	\end{subfigure}
\hfill
	\begin{subfigure}{0.495\linewidth}
		\includegraphics[width=\linewidth]{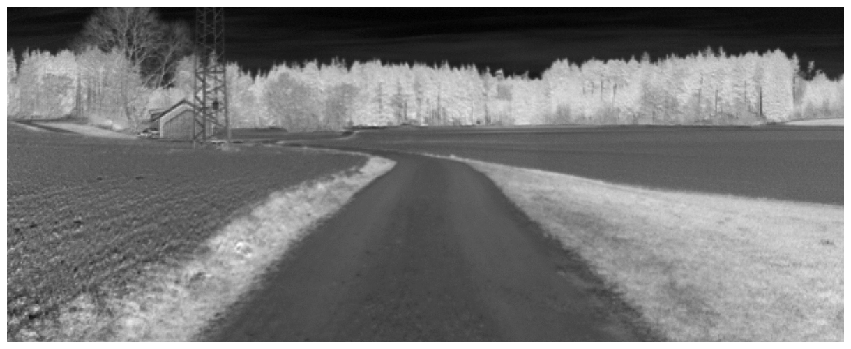}
		\caption*{NDVI}
	\end{subfigure}
	\begin{subfigure}{0.495\linewidth}
		\includegraphics[width=\linewidth]{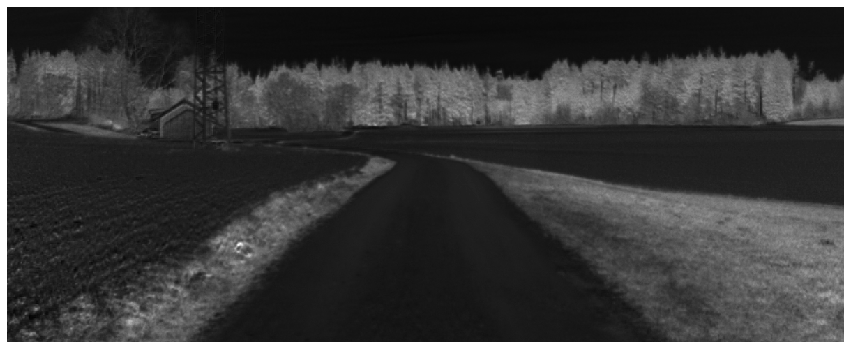}
		\caption*{EVI}
	\end{subfigure}
	\caption{Both vegetation indices NDVI and EVI make use of the high reflectivity of vegetation in the near infrared spectrum and the low reflectivity of vegetation in the visible red spectrum.}
	\label{fig:vegetation-indices}
\end{figure}

\noindent The appearance of spatial content in NIR imagery differs in a few ways from their appearance in the visible spectrum (see \cref{fig:vegetation-indices}). 
For vegetation in particular, where the chlorophyll in the vegetation appears transparent in the NIR spectrum, allowing the light in the NIR spectrum to reflect on the water contained in the vegetation~\cite{pemo:molecularstructure_nassau2001}. 
The intensity of the NIR reflection depends on the season and the type of plants, which is why vegetation indices tend to combine the NIR response with other image properties from the VIS spectrum. 
For instance, the common Normalized Difference Vegetation Index (NDVI) metric makes use of the low reflectivity of vegetation in the red channel $\red$ of the debayered VIS image in combination with the high reflectivity in the NIR image.
\begin{equation}
\ndvi = 
\begin{cases}
\scaleto{\frac{\nir - \red}{\nir + \red}}{18pt} & \quad\quad \text{otw.}\\
0 & \quad\text{ if } \nir = \red = 0
\end{cases}
\label{eq:ndvi}
\end{equation} 
Both the NIR and VIS image require the same bit depth for the NDVI calculation. The result NDVI value is in $\lbrack-1,1\rbrack$ range. 
We can observe that $\ndvi \geq 0$ for pixels, where the reflectance in the NIR spectrum is higher than the reflectance in the R channel and vice versa.
As observed in remote sensing applications, this translates to negative NDVI values for bodies of water and NDVI values close to zero for surface types like rocks, sands and concrete surfaces.
Positive NDVI values come up for vegetation like crops, shrubs, grasses and forests \cite{pemo:ndviproperties_jones2010}. \\
In our experiments, we will also take a closer look at the Enhanced Vegetation Index. The EVI was introduced to compensate soil and atmospheric effects and incorporates the light from the visible blue channel $\blue$ as well. We had troubles finding a consistent formula of the EVI index, we therefore rely on the following formula~\cite{pemo:evi_huete1999modis}. 
\begin{equation}
\evi = 
\begin{cases}
\scaleto{\frac{2 \cdot (\nir - \red)}{\nir +~C_{1} \cdot \red +~C_{2} \cdot \blue}}{18pt} & \quad \text{otw.} \\
 & \text{if } \nir = \red \\ \quad 0 & = \blue = 0 
\end{cases}
\label{eq:evi}
\end{equation}
\begin{equation*}
\text{where } C_{1} = 6.0 \text{ and } C_2 = 7.5 
\end{equation*}
The EVI index is in $\lbrack-\frac{2}{C_{1}},2\rbrack$ range.
The weights $C_{1}$ and $C_{2}$ adjust the use of the blue channel in aerosol correction of the red channel. 
The mentioned atmospheric effects do not relate to the image acquisition process from the ground with a camera system, but we will use the \cref{eq:evi} for the sake of consistency. 
In \cref{fig:vegetation-indices} we present the NDVI and EVI image for a scene from the \dataset. 

\begin{figure*}[ht!]
	\hrulefill
	
	\vspace{.5em}
	\begin{subfigure}{0.24\linewidth}
		\begin{center}
			\small VIS
			\vspace{0.5em}
		\end{center}
		\includegraphics[width=\linewidth]{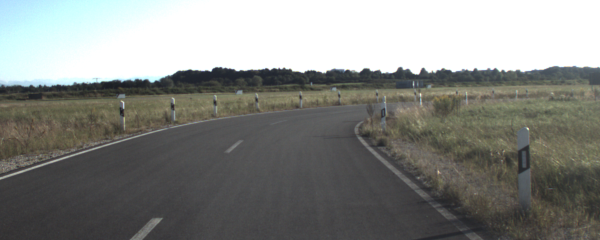}
	\end{subfigure}
	\begin{subfigure}{0.24\linewidth}
		\begin{center}
			\small NIR
			\vspace{0.5em}
		\end{center}
		\includegraphics[width=\linewidth]{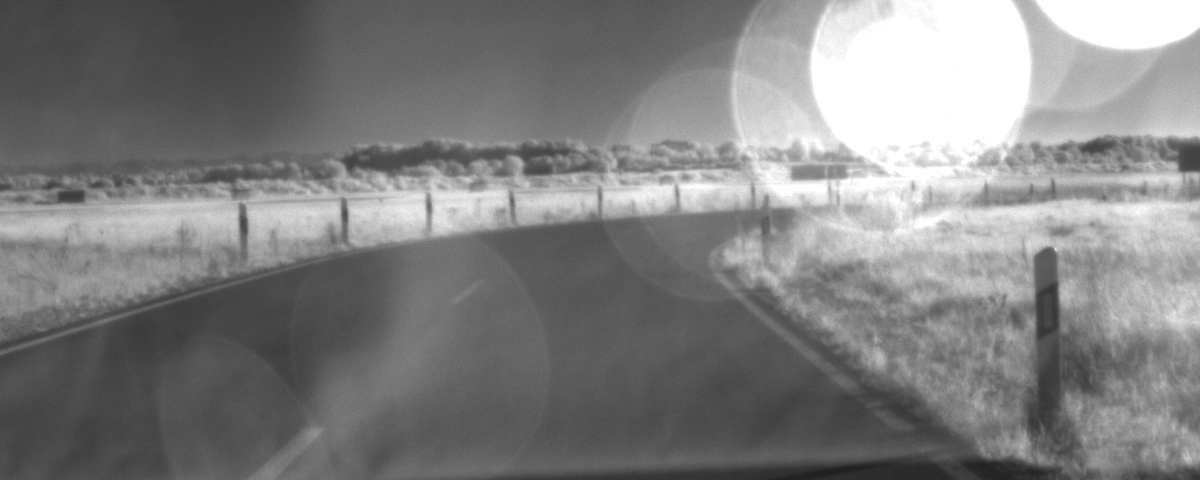}
	\end{subfigure}
	\begin{subfigure}{0.24\linewidth}
		\begin{center}
			\small Prediction
			\vspace{.5em}
		\end{center}
		\includegraphics[width=\linewidth]{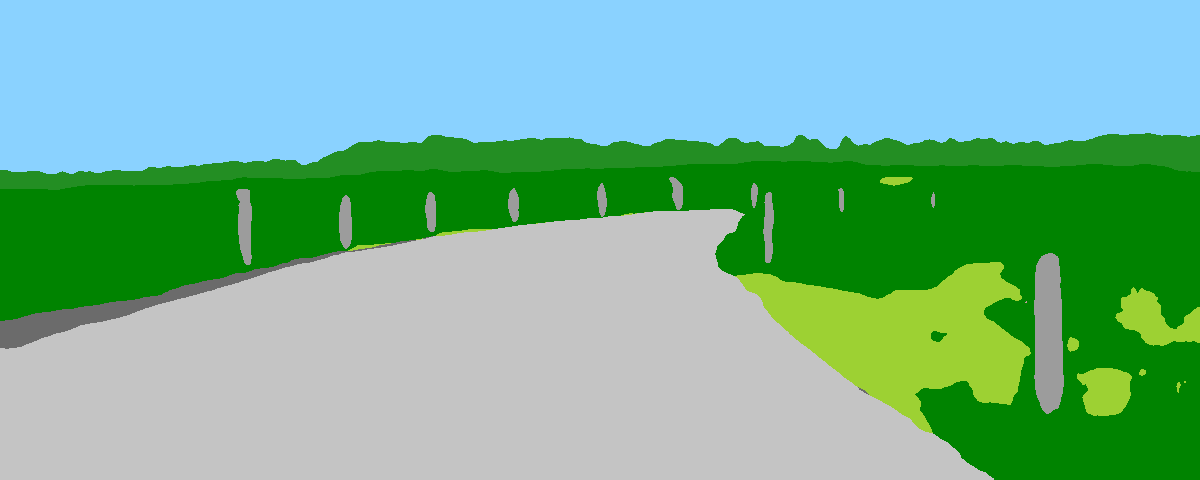}
	\end{subfigure}
	\begin{subfigure}{0.24\linewidth}
		\begin{center}
			\small Ground Truth
			\vspace{.5em}
		\end{center}
		\includegraphics[width=\linewidth]{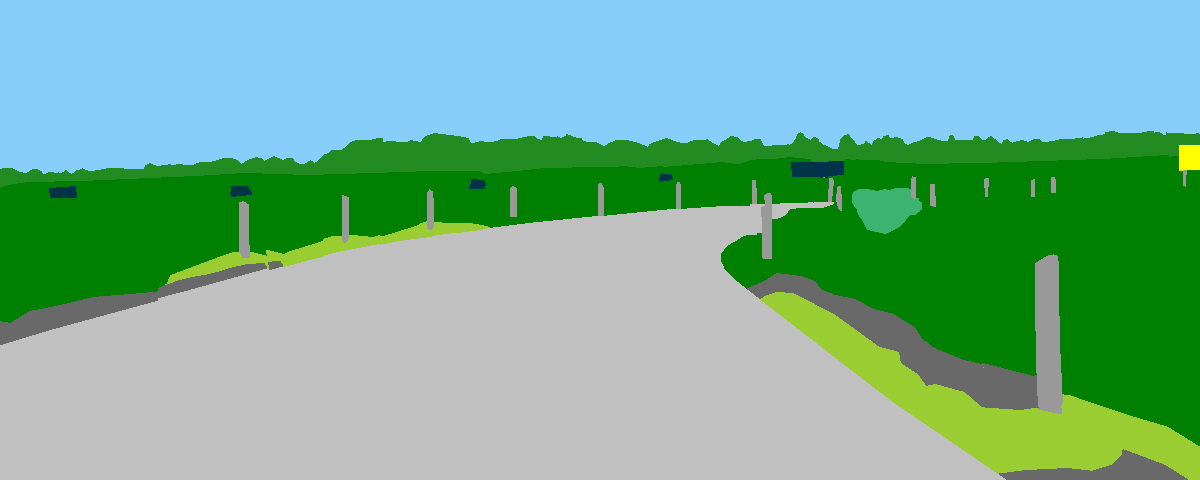}
	\end{subfigure}

	\hfill

	\begin{subfigure}{0.24\linewidth}
		\vspace{.5em}
		\includegraphics[width=\linewidth]{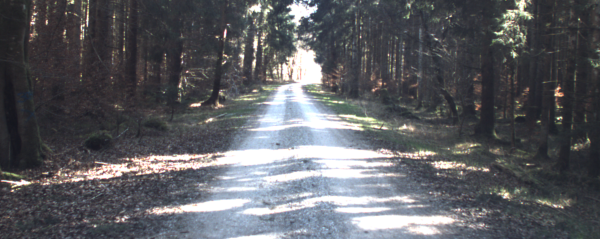}
	\end{subfigure}
	\begin{subfigure}{0.24\linewidth}
		\vspace{.5em}
		\includegraphics[width=\linewidth]{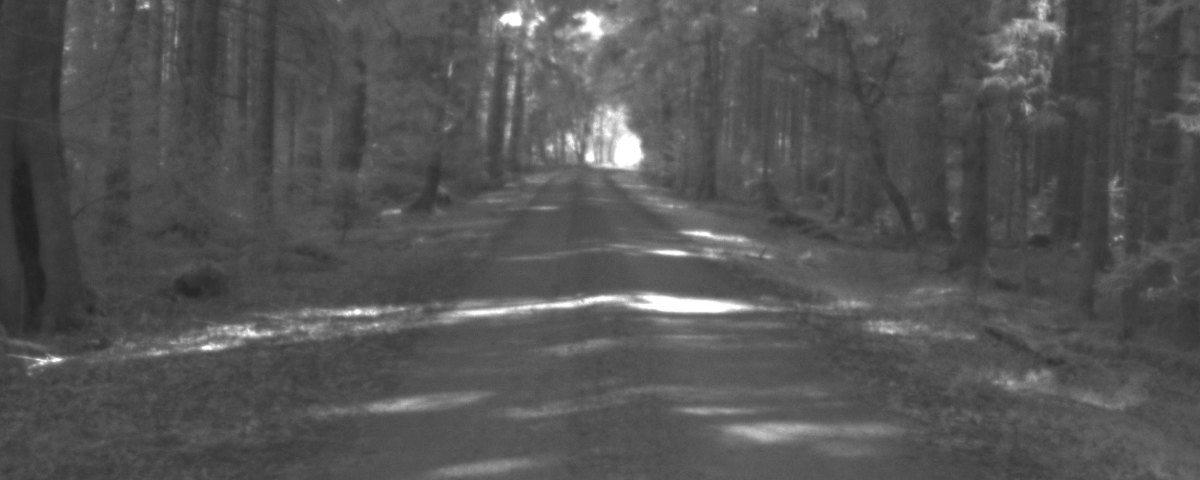}
	\end{subfigure}
	\begin{subfigure}{0.24\linewidth}
		\vspace{.5em}
		\includegraphics[width=\linewidth]{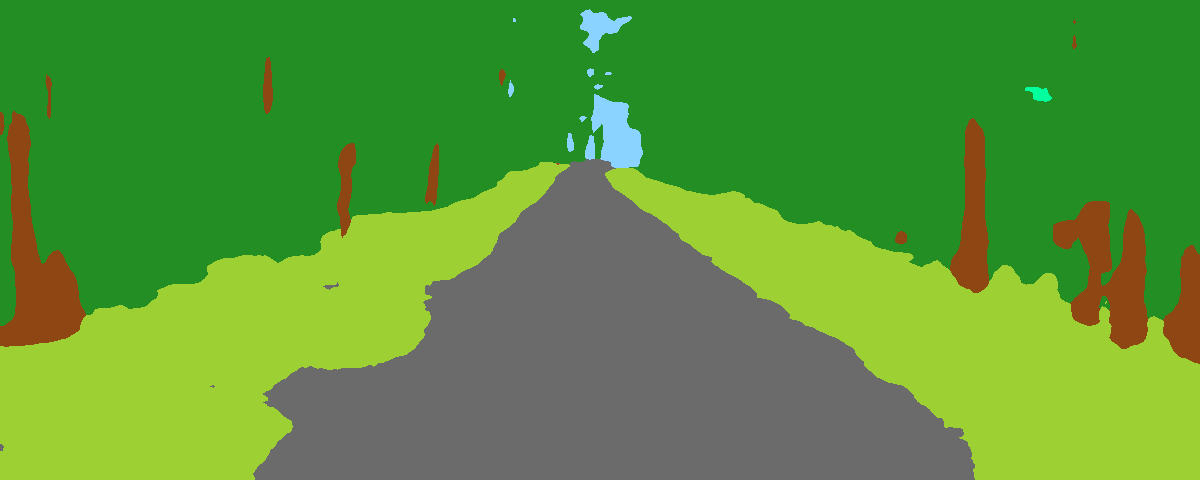}
	\end{subfigure}
	\begin{subfigure}{0.24\linewidth}
		\vspace{.5em}
		\includegraphics[width=\linewidth]{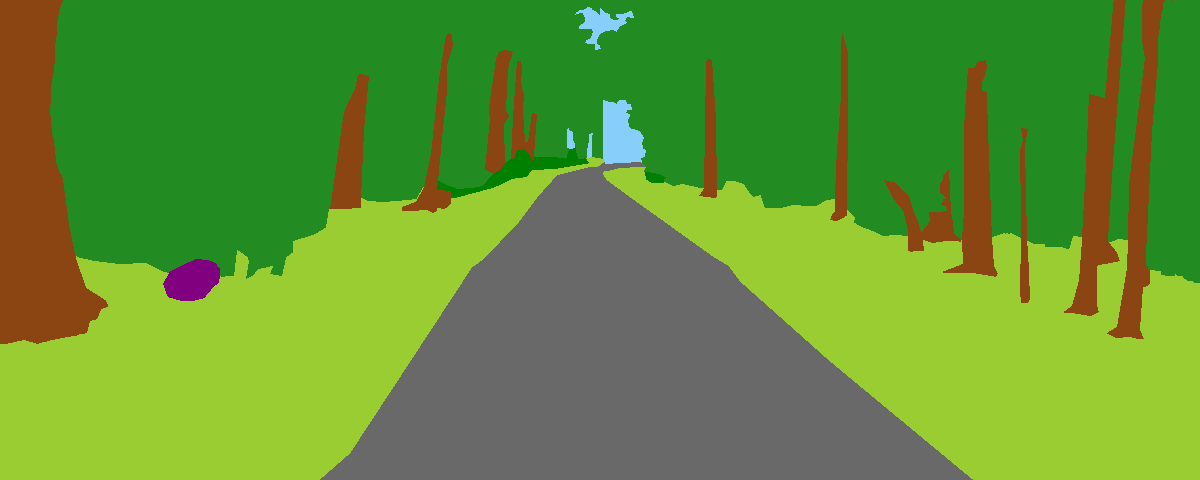}
	\end{subfigure}

	\hfill

	\begin{subfigure}{0.24\linewidth}
		\vspace{.5em}
		\includegraphics[width=\linewidth]{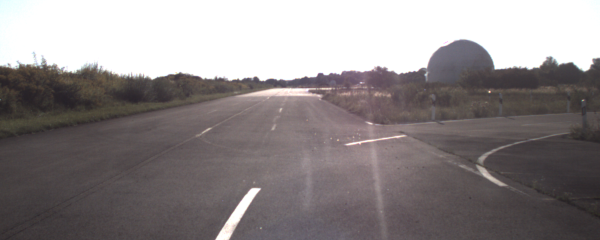}
	\end{subfigure}
	\begin{subfigure}{0.24\linewidth}
		\vspace{.5em}
		\includegraphics[width=\linewidth]{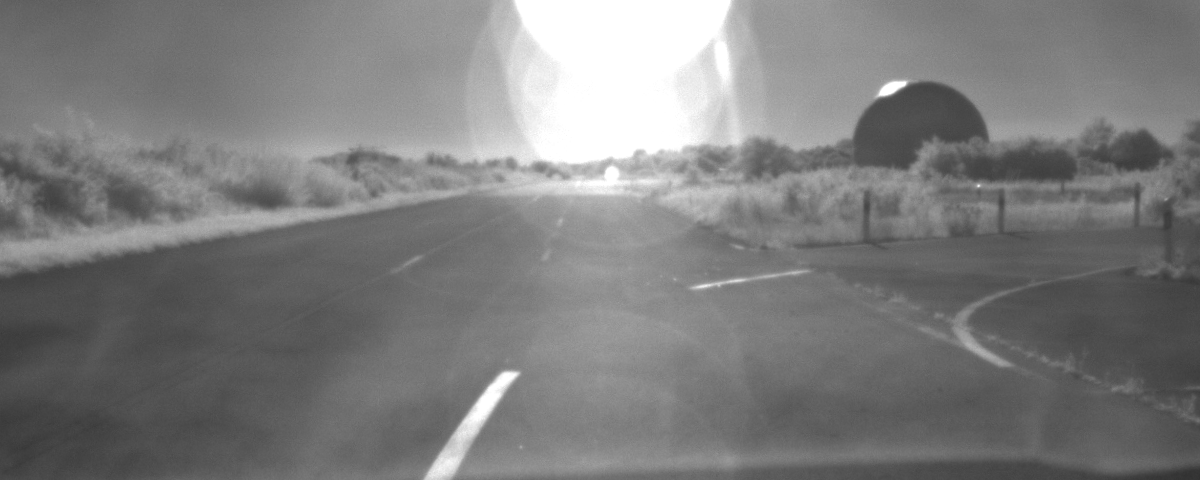}
	\end{subfigure}
	\begin{subfigure}{0.24\linewidth}
		\vspace{.5em}
		\includegraphics[width=\linewidth]{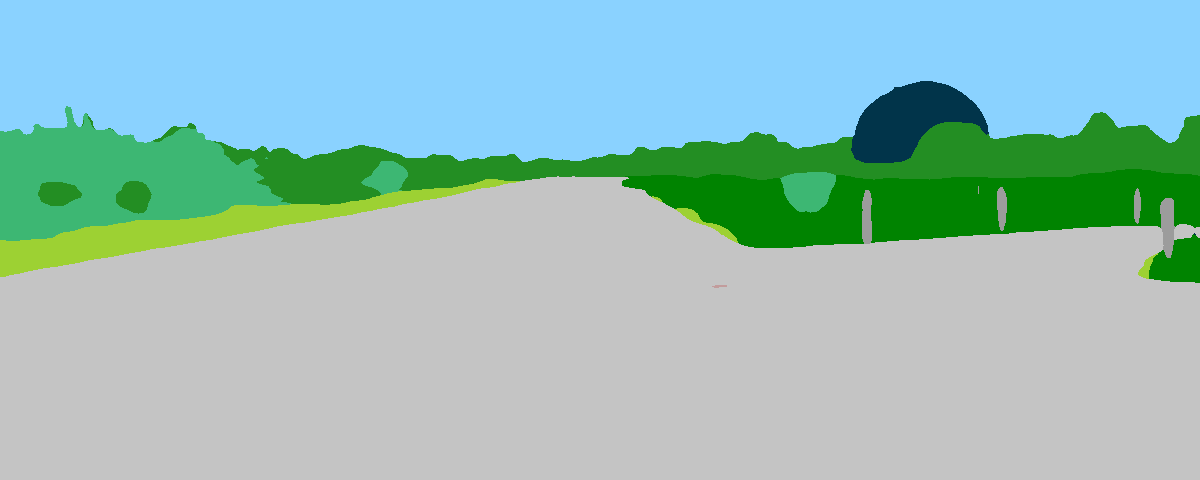}
	\end{subfigure}
	\begin{subfigure}{0.24\linewidth}
		\vspace{.5em}
		\includegraphics[width=\linewidth]{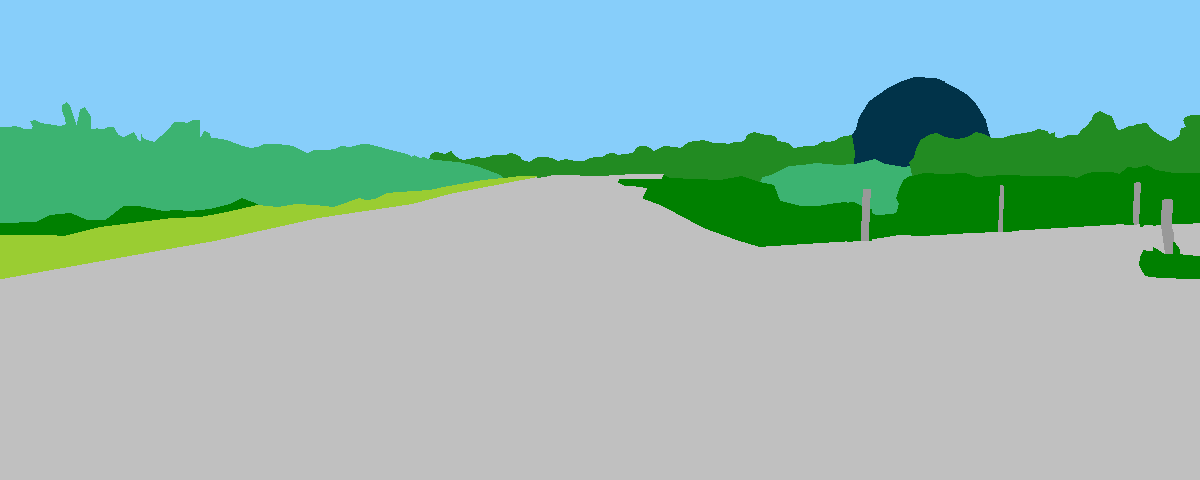}
	\end{subfigure}

	\hfill

	\begin{subfigure}{0.24\linewidth}
		\vspace{.5em}
		\includegraphics[width=\linewidth]{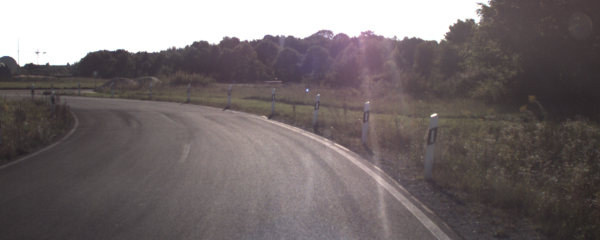}
	\end{subfigure}
	\begin{subfigure}{0.24\linewidth}
		\vspace{.5em}
		\includegraphics[width=\linewidth]{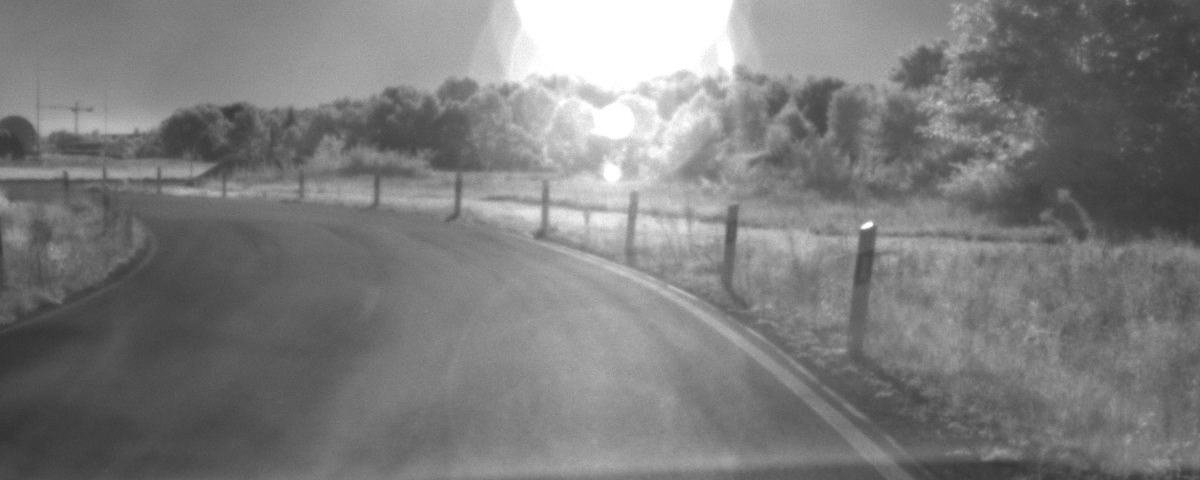}
	\end{subfigure}
	\begin{subfigure}{0.24\linewidth}
		\vspace{.5em}
		\includegraphics[width=\linewidth]{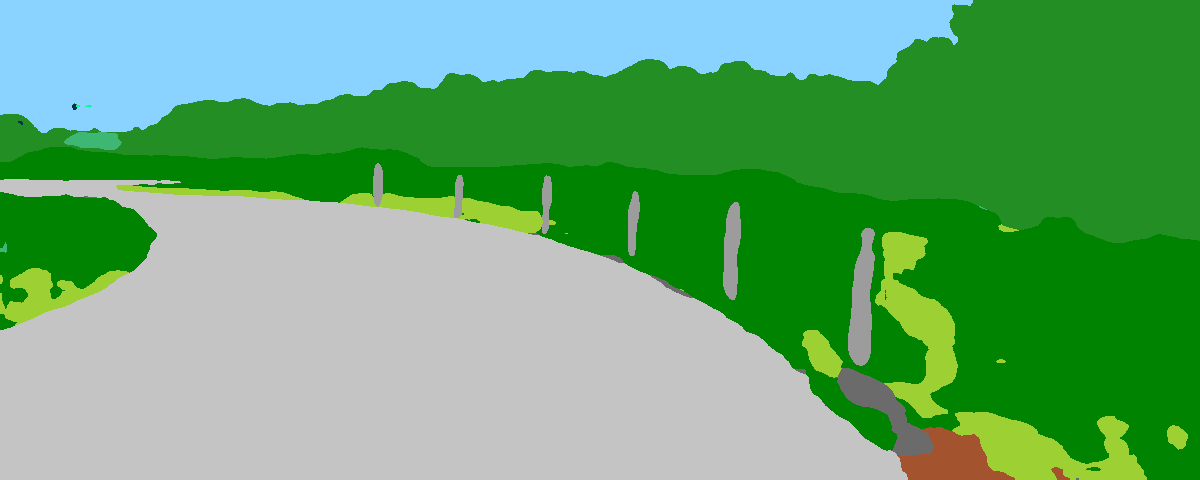}
	\end{subfigure}
	\begin{subfigure}{0.24\linewidth}
		\vspace{.5em}
		\includegraphics[width=\linewidth]{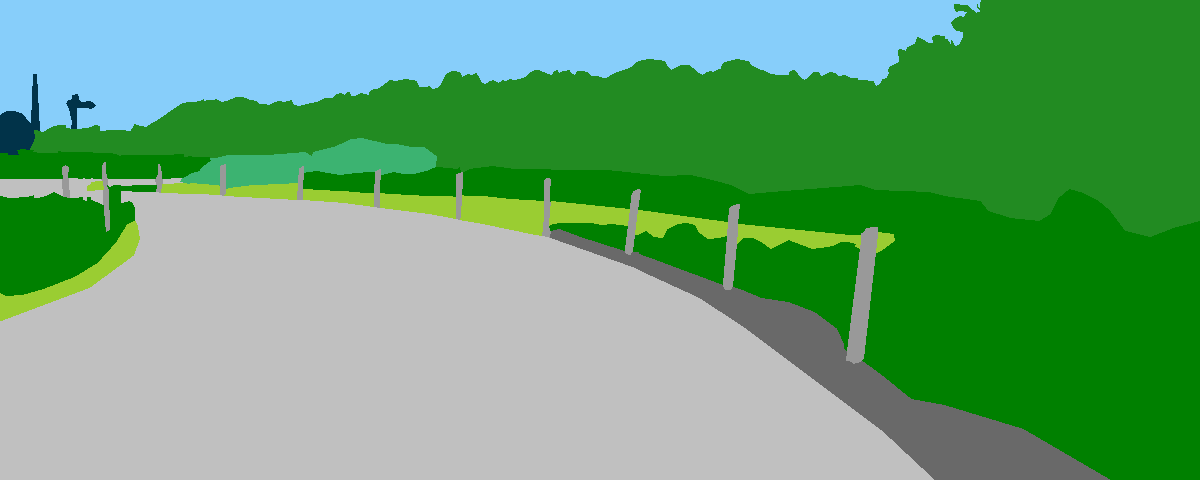}
	\end{subfigure}

	\hfill

	\begin{subfigure}{0.24\linewidth}
		\vspace{.5em}
		\includegraphics[width=\linewidth]{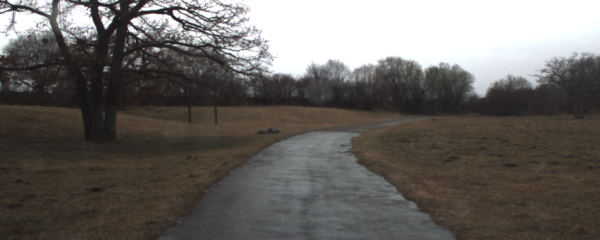}
	\end{subfigure}
	\begin{subfigure}{0.24\linewidth}
		\vspace{.5em}
		\includegraphics[width=\linewidth]{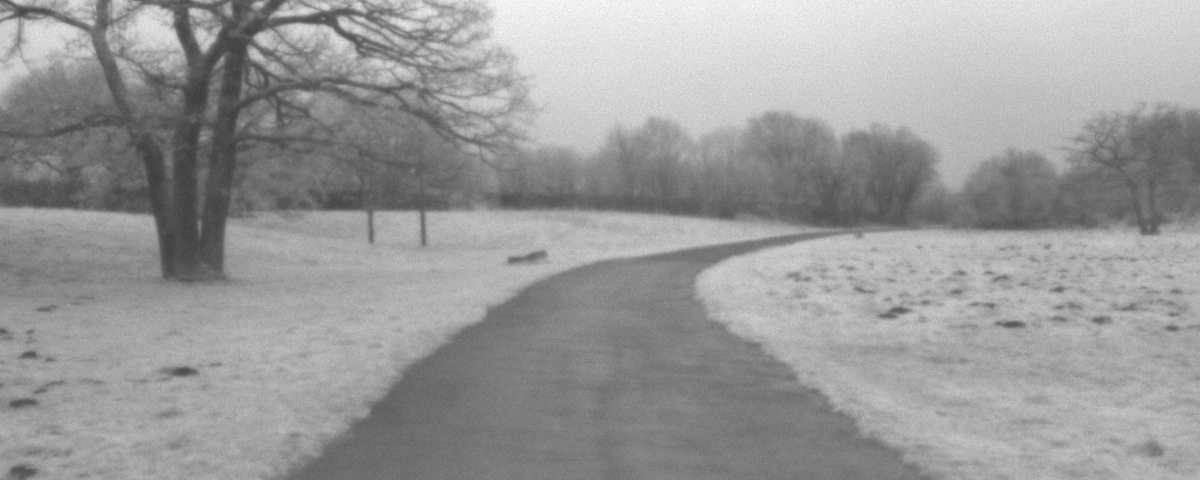}
	\end{subfigure}
	\begin{subfigure}{0.24\linewidth}
		\vspace{.5em}
		\includegraphics[width=\linewidth]{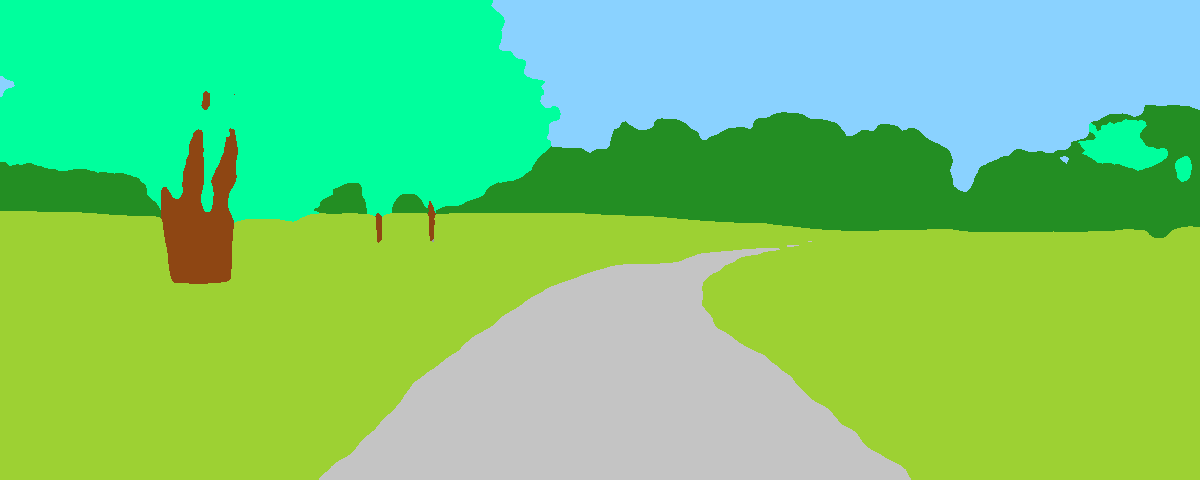}
	\end{subfigure}
	\begin{subfigure}{0.24\linewidth}
		\vspace{.5em}
		\includegraphics[width=\linewidth]{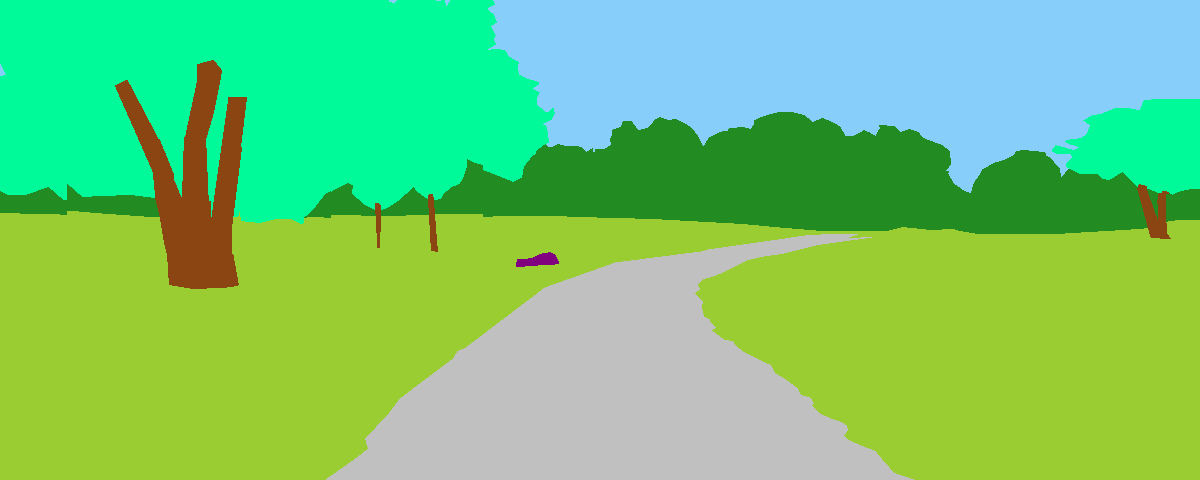}
	\end{subfigure}
	\vspace{1em}
	
	\hrulefill
	\caption{
		A qualitative evaluation of the DeepLabV3+ prediction performance with different types of vegetation, such as a \textcolor{tree_trunk}{tree trunk}, \textcolor{tree_crown}{tree crown}, \textcolor{low_grass}{low grass}, \textcolor{high_grass}{high grass}, \textcolor{forest}{forest} and \textcolor{bush}{bush}. The presented images also show the \textcolor{pole}{pole} and \textcolor{obstacle}{obstacle} class, which we didn't quantitatively evaluate in \cref{tab:tas-nir-experiment}.
	}
	\label{fig:qualitative-evaluation}
\end{figure*}
\begin{table*}[htpb!]
	\caption{The Intersection over Union (IoU) in percent for each semantic class of interest and the mean Intersection over Union (mIoU) over all semantic classes of interest. We use local temperature scaling (LTS) to calibrate the output of DeepLabV3+. The addition of a image histogram-based approach alone ($\beta=0.75$) shows no significant improvement. We observe a significant improvement from the fully-connected conditional random field (CRF with $\theta_{\alpha} = 10$, $\theta_{\beta} = 13$ ). \Checkmark$_{x}$ notes, that the CRF uses image modality $x$ as structural input during inference.}
	\label{tab:tas-nir-experiment}
	\vspace{3em}
	\centering
	\begin{tabular}{*{12}{c|c|c|c|c|ccccccccc}}
		
		network & LTS & Histogram & CRF & mIoU & 
		\begin{rotate}{45}\small{
				asphalt \textcolor{asphalt}{$\blacksquare$}
		}\end{rotate} & 
		\begin{rotate}{45}\small{
				gravel \textcolor{gravel}{$\blacksquare$}
		}\end{rotate} & 
		\begin{rotate}{45}\small{
				soil \textcolor{soil}{$\blacksquare$}
		}\end{rotate} & 
		\begin{rotate}{45}\small{
				low grass \textcolor{low_grass}{$\blacksquare$}
		}\end{rotate} & 
		\begin{rotate}{45}\small{
				high grass \textcolor{high_grass}{$\blacksquare$}
		}\end{rotate} & 
		\begin{rotate}{45}\small{
				bush \textcolor{bush}{$\blacksquare$}
		}\end{rotate} & 
		\begin{rotate}{45}\small{
				tree crown \textcolor{tree_crown}{$\blacksquare$}
		}\end{rotate} & 
		\begin{rotate}{45}\small{
				tree trunk \textcolor{tree_trunk}{$\blacksquare$}
		}\end{rotate} & 
		\begin{rotate}{45}\small{
				forest \textcolor{forest}{$\blacksquare$}
		}\end{rotate}  \\
		\cline{1-14}
		& 
		\xmark & 
		\xmark & 
		\xmark & 
		\small{30.61} & 
		\small{50.24} & 
		\small{34.14} & 
		\small{2.55} & 
		\small{56.86} & 
		\small{30.98} & 
		\small{10.14} &
		\small{7.43} & 
		\small{10.80} & 
		\small{72.38} 
		\\
		\cline{2-14}
		& 
		\Checkmark & 
		NDVI & 
		\xmark & 
		\small{30.56} & 
		\small{50.08} & 
		\small{34.17} & 
		\small{2.50} & 
		\small{56.75} & 
		\small{30.80} & 
		\small{10.26} &
		\small{7.42} & 
		\small{10.74} & 
		\small{72.39} 
		\\
		\cline{2-14}
		& 
		\Checkmark & 
		EVI & 
		\xmark & 
		\small{30.62} & 
		\small{\textbf{50.31}} & 
		\small{34.13} & 
		\small{2.54} & 
		\small{56.88} & 
		\small{30.97} & 
		\small{10.13} & 
		\small{7.44} & 
		\small{10.76} & 
		\small{72.40} 
		\\
		\cline{2-14}
		& 
		\Checkmark & 
		NDVI & 
		~\Checkmark$_{\textrm{NDVI}}$ & 
		\small{47.97} & 
		\small{50.16} & 
		\small{61.02} & 
		\small{23.11} & 
		\small{\textbf{62.48}} & 
		\small{46.93} & 
		\small{37.78} &
		\small{46.34} & 
		\small{30.65} & 
		\small{\textbf{73.12}} 
		\\
		\cline{2-14}
		& 
		\Checkmark & 
		EVI & 
		\Checkmark$_{\textrm{EVI}}$ & 
		\small{50.16} & 
		\small{47.27} & 
		\small{62.08} & 
		\small{30.40} & 
		\small{62.17} & 
		\small{\textbf{48.43}} & 
		\small{38.59} &
		\small{56.98} & 
		\small{\textbf{32.64}} & 
		\small{72.88} 
		\\
		\cline{2-14}
		\multirow{3}{*}{\begin{rotate}{90}\small{~~DeepLabv3+}\end{rotate}} & 
		\Checkmark & 
		\xmark & 
		\Checkmark$_{\textrm{VIS}}$ & 
		\small{\textbf{52.19}} & 
		\small{45.31} & 
		\small{70.50} & 
		\small{\textbf{43.10}} & 
		\small{60.09} & 
		\small{47.99} & 
		\small{\textbf{41.08}} &
		\small{\textbf{59.89}} & 
		\small{29.23} & 
		\small{72.60} 
		\\
		
	\end{tabular}
\end{table*}

\subsection{Late-fusion of VIS+NIR Predictions}
\noindent Many instances of the same semantic class share similar values on both vegetation indices.
The vegetation classes with a similar visual appearance in the VIS image can be discriminated among the semantic class in the NIR image.
A lack of semantically segmented VIS+NIR image pairs prevents us from training a multimodal network for this domain.
We therefore suggest adding a post-processing stage to the VIS-only neural network output by calibrating its outputs and adjusting the predictions based on the vegetation index values for each pixel position of the prediction.
The prediction using the vegetation index is based on the accumulated image histograms of each semantic class across the validation set. \\
The image histogram for each semantic class in the NDVI image is clustered into 16 bins, while the histograms for the EVI images consist of 20 bins. 
The weight of a bin is defined as the ratio of the semantic pixels in a bin compared to all semantic pixels in the histogram of a semantic class. \\
The calibrated prediction for each pixel is supplemented with the histogram-based prediction of the respective NDVI and EVI pixel by adding the normalized bin weights $\omega(x)$ for each semantic class $k$.
\begin{equation}
\hat{\mathbb{Q}}_{n}(x, T_{n}, \omega) = \max_{k \in K} \beta~\omega(x)^{(k)} + \sigma_{\mathrm{SM}}(\frac{\mathbf{z}_{n}(x)}{T_{n}(x)})^{(k)}
\label{eq:fusion}
\end{equation}

The hyperparameter $\beta$ weights the influence of the histogram-based predictions $\omega(x)$. 
$\beta = 0.75$ is set for both the NDVI and EVI image predictions and $\beta$ has been optimized on the validation split.

To smooth the fused predictions, we use a fully connected conditional random field \cite{crf}. The possible labelings of an image conditioned over the input image pixel intensities are characterized by a graph and its cliques. The cliques induce a potential, which assign a cost to assigning labels to neighboring pixels \cite{kernel_crf}. The potentials consists of two kernels. 

The first is a so-called smoothness kernel, that removes small isolated regions in the labeling \cite{smoothness_kernel_crf}. 
The second kernel is the appearance kernel, which penalizes if nearby pixels have different semantic labels (\textit{nearness}) and penalizes if pixels with similar pixel intensities in the input image have different semantic labels in the prediction (\textit{similarity}). 

A higher correlation between the pixel intensities in the vegetation indices to the semantic classes of different surface and vegetation types can be observed.
We investigate the effect of passing the vegetation indices or the NIR image to the CRF as input image improves the final semantic segmentation. 

\section{Experiment}
\label{sec:experiment}
\noindent We quantitatively evaluate our method by comparing the prediction performance of a semantic segmentation model, that only makes use of the VIS images from the test split, to our combined approach of both calibrated neural network predictions on VIS data and hand-crafted VIS+NIR vegetation indices. \\
For our experiment we have trained a DeepLabv3+ network \cite{pemo:deeplab_chen2018eccv} with the \numtrain VIS images of the TAS500 \cite{pemo:tas500_metzger2020icpr} dataset. 
The images in the training dataset are cropped to $1200~\times$ \SI{480}{\px} and are matched to the same region of interest as in the \dataset. 
The neural network has been trained for 80.000 iterations in total. \\
In \cref{tab:tas-nir-experiment} we compare the general semantic segmentation performance of all methods and compare the performance on a per-class level.
The low IoU scores for specific semantic classes like \textcolor{bush}{bush}, \textcolor{soil}{soil}, \textcolor{tree_trunk}{tree trunk} and \textcolor{tree_crown}{tree crown} in \cref{tab:tas-nir-experiment} is caused by only very few occurrences in the \dataset test split. This leads to misclassifications of these few instances to more heavily effect the IoU score. In \cref{fig:qualitative-evaluation} we present some predictions of our proposed method. 

\begin{figure}[h!]
	\begin{subfigure}{0.495\linewidth}
		\includegraphics[width=\linewidth]{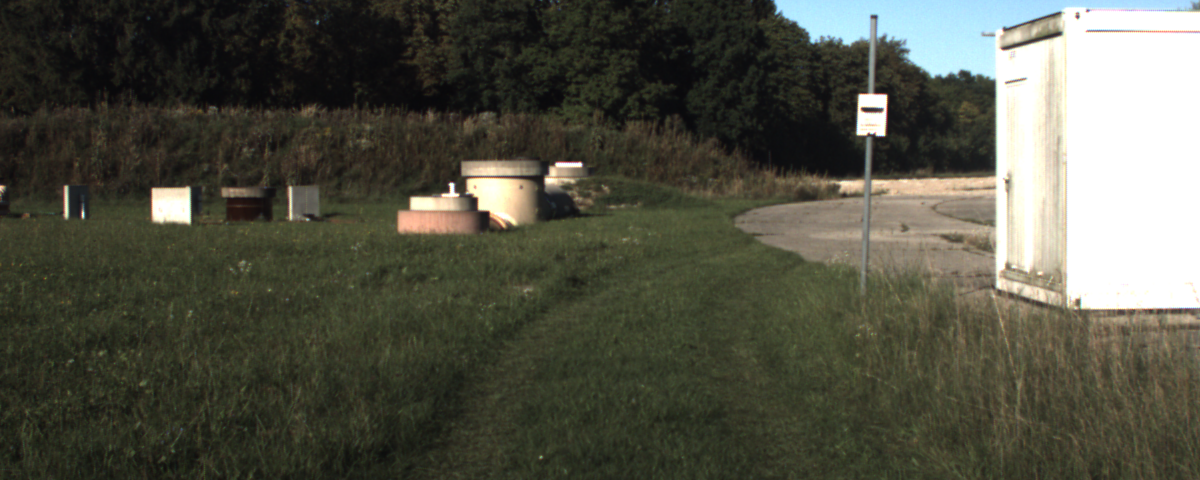}
		\caption*{VIS}
	\end{subfigure}
	\begin{subfigure}{0.495\linewidth}
		\includegraphics[width=\linewidth]{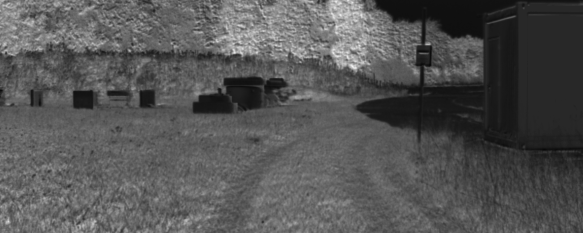}
		\caption*{EVI}
	\end{subfigure}
	\hfill
	\begin{subfigure}{0.495\linewidth}
		\includegraphics[width=\linewidth]{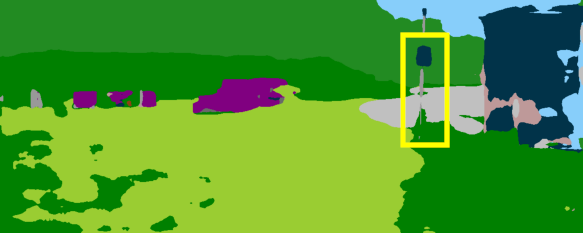}
		\caption*{DeepLabv3+}
	\end{subfigure}
	\begin{subfigure}{0.495\linewidth}
		\includegraphics[width=\linewidth]{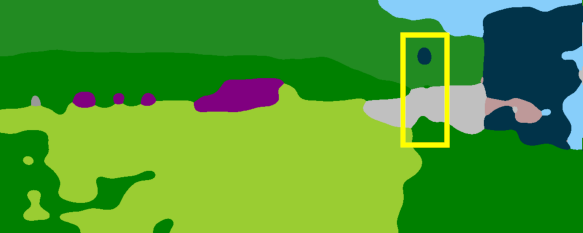}
		\caption*{CRF on EVI}
	\end{subfigure}
	\caption{For the conditional random field (CRF) we can observe how the smoothness kernel can lead to thin structures like the \textcolor{pole}{pole} in the yellow box (\textcolor[RGB]{255,255,0}{$\blacksquare$}) to disappear in the final segmentation.}
	\label{fig:failure-mode}
\end{figure}

\section{Conclusions}

\noindent In this work, we introduced the novel \dataset for the fine-grained semantic segmentation of ground surface and vegetation types in unstructured outdoor environments from VIS+NIR image pairs.
The image histogram-based approach to supplement the semantic segmentation coming from the VIS image has shown only very little influence on improving the overall semantic segmentation performance. 
The use of a CRF improves the performance significantly. 
The CRF can fill small semantic regions in the final prediction.
This comes at the cost of missing thin obstacles like the pole in \cref{fig:failure-mode}. 


\end{document}